\documentclass[journal,onecolumn]{IEEEtran}
\ifCLASSINFOpdf
\else
\fi
\usepackage{epsfig}
\usepackage{graphicx}
\usepackage{amsmath}
\usepackage{amssymb}
\usepackage{commath}
\usepackage{bbding}
\usepackage{cite}
\usepackage{cleveref}
\usepackage{booktabs,multirow}

\begin{document}
%
\title{Super-Resolution of Wavelet-Encoded Images}
%
%
%

\author{Vildan~Atalay~Aydin and~Hassan~Foroosh
\thanks{Vildan Atalay Aydin and Hassan Foroosh are with the Department of Computer Science, University of Central Florida, Orlando,
FL, 32816 USA (e-mails: vatalay@knights.ucf.edu and foroosh@cs.ucf.edu).}
}

\maketitle

\begin{abstract}
Multiview super-resolution image reconstruction (SRIR) is often cast as a resampling problem by merging non-redundant data from multiple low-resolution (LR) images on a finer high-resolution (HR) grid, while inverting the effect of the camera point spread function (PSF). One main problem with multiview methods is that resampling from nonuniform samples (provided by LR images) and the inversion of the PSF are highly nonlinear and ill-posed problems. Non-linearity and ill-posedness are typically overcome by linearization and regularization, often through an iterative optimization process, which essentially trade off the very same information (i.e. high frequency) that we want to recover. We propose a novel point of view for multiview SRIR: Unlike existing multiview methods that reconstruct the entire spectrum of the HR image from the multiple given LR images, we derive explicit expressions that show how the high-frequency spectra of the unknown HR image are related to the spectra of the LR images. Therefore, by taking any of the LR images as the reference to represent the low-frequency spectra of the HR image, one can reconstruct the super-resolution image by focusing only on the reconstruction of the high-frequency spectra. This is very much like single-image methods, which extrapolate the spectrum of one image, except that we rely on information provided by all other views, rather than by prior constraints as in single-image methods (which may not be an accurate source of information). This is made possible by deriving explicit closed-form expressions that define how the local high frequency information that we aim to recover for the reference high resolution image is related to the local low frequency information in the sequence of views. The locality of these expressions due to modeling using wavelets reduces the problem to an exact and linear set of equations that are well-posed and solved algebraically without requiring regularization or interpolation. Results and comparisons with recently published state-of-the-art methods show the superiority of the proposed solution.
\end{abstract}


\IEEEpeerreviewmaketitle

\section{Introduction} \label{sec:intro}
Super-resolution \cite{Foroosh_Chellappa_1999,Foroosh_etal_1996,shekarforoush19953d,lorette1997super,shekarforoush1998multi,berthod1994refining,shekarforoush1999conditioning,jain2008super,shekarforoush1999super} has a wide-range of applications in various areas of imaging and computer vision, such as self-localization \cite{Junejo_etal_2010,junejo2008gps,Cao_Foroosh_2007}, image annotation  \cite{Tariq_etal_2017_2,Tariq_etal_2017,tariq2013exploiting,tariq2015feature,tariq2014scene,tariq2015t}, surveillance \cite{Junejo_etal_2007,Junejo_Foroosh_2008,junejo2007trajectory}, action recognition \cite{Sun_etal_2015,Ashraf_etal_2014,Ashraf_etal_2013,Shen_Foroosh_2009,shen2008view,sun2011action,ashraf2014view,shen2008action,ashraf2010view,boyraz122014action,sun2014feature,ashraf2012human}, target tracking \cite{Shu_etal_2016,shekarforoush2000multi,Milikan_etal_2017,millikan2015initialized}, shape description and object recognition \cite{Zhang_2015,Cakmakci_etal_2008,Cakmakci_etal_2008_2,ali2016character,ali2016character}, image-based rendering \cite{alnasser2006image,balci2006real,balci2006image,shen2006video}, and camera motion estimation \cite{Junejo_etal_2011,Cao_Foroosh_2007,Cao_Foroosh_2006,cao2004camera,junejo2006calibrating,cao2006self,cao2006camera,ashraf2007near}, to name a few. The goal of multiview super-resolution image reconstruction is to obtain a high resolution (HR) image by fusing a sequence of degraded or aliased low resolution (LR) images of the same scene where degradation can be a consequence of motion, camera optics, atmosphere, insufficient sampling, etc. 
\\

Approaches to solve the SRIR problem can be classified into frequency domain, interpolation, regularization, and learning-based methods (\cite{park2003super,tian2011survey}). 

Fourier-based methods make use of the aliasing property of LR images in order to reconstruct an HR image. Even though these methods are intuitive and have low computational complexity, due to their global nature, they only allow linear space invariant blur (PSF). Moreover, it is difficult to identify a global frequency-domain {\em a priori} knowledge to overcome ill-posedness. Some examples of Fourier domain techniques include the following works: \cite{tsai} exploit the relationship between Continuous Fourier Transform (CFT) of the unknown HR scene and Discrete Fourier Transform (DFT) of the shifted and sampled LR images; \cite{rivenson2010single} utilize double random phase encoding in the imaging process in order to achieve SRIR with a single image. \cite{robinson2010efficient} apply combined Fourier-wavelet de-convolution and de-noising algorithm; and \cite{vandewalle2007super} perform registration and reconstruction in the Fourier domain using multiple unregistered images.

Spatial-domain interpolation-based methods (\cite{irani1991improving,zhou2012interpolation,zomet2001robust}), on the other hand, tackle Fourier-domain obstacles by fusing the information from all LR images using a general interpolation technique, such as the nearest neighbor, bilinear, bicubic, etc. However, these methods result in overly smoothed images. As an example of interpolation based methods, \cite{irani1991improving} update the HR estimate by iteratively back-projecting the difference between the approximation and exact image. Moreover, \cite{zhou2012interpolation} utilize multi-surface fitting; whereas, \cite{de2013edge} interpolate along the edge direction. \cite{shekarforoush1999data} generalize Papoulis’ sampling theorem to merge nonuniform samples of multiple channels.

In order to stabilize the ill-posed problem of SRIR, regularization-based methods optimize a cost function with a regularization term by incorporating prior knowledge. Some of these methods employ probabilistic estimators such as the maximum likelihood (\cite{kim2013fast,elad2001fast}), maximum a posteriori (\cite{shen2007map,belekos2010maximum}), and Bayesian (\cite{babacan2011variational,bishop2006bayesian,pickup2007overcoming}). The problems with these methods are determination of the prior model, high computational cost, and over-smoothing caused by regularization. More examples of regularization-based methods include the following works: \cite{marquina2008image} employ Bregman iteration for Total Variation Regularization; \cite{huang2010super} carry out canonical correlation analysis for human face SRIR task; \cite{farsiu2004fast} minimize $L_1$ norm and utilize regularization based on a bilateral prior; \cite{farsiu2006multiframe} minimize a multi-term cost function; and \cite{yuan2012multiframe} constrain the SRIR process by using a spatially weighted TV model for different image regions. \cite{lorette1997super} investigate the contradiction between multichannel super resolution and regularization within the adaptive regularization framework.

Finally, machine learning is also used for SRIR (\cite{elad2009example,freeman2002example,yang2010image}), where an HR image is obtained from LR images by utilizing training sets of LR/HR images or patch pairs. The problems with these methods again include their high computational cost, and correspondence ambiguities between HR and LR images. \cite{wu2011learning} utilize kernel partial least squares to implement regression along with a compensation of the primitive HR image with a residual HR image. Clustering and supervised neighbor embedding is employed by \cite{zhang2011partially}. \cite{glasner2009super} combine multi-image SRIR and example-based approaches based on the assumption that patches in natural images recur many times inside the image; \cite{kappeler2016video} propose a video super resolution method with a convolutional neural network (CNN) that is trained on both spatial and temporal dimensions of videos; and \cite{dai2017sparse} extend video bilevel dictionary learning to multiframe super resolution using motion estimation.

In order to overcome the drawbacks of aforementioned methods, recent research in SRIR explores wavelet-based techniques (\cite{ji2009robust, robinson2010efficient, demirel2011image, dong2011image, tong2007super}). The intuition behind these approaches is that the LR images can be used to model the low-pass subbands of the unknown HR images, in order to reconstruct the high frequency information lost during image acquisition. The use of wavelets is also motivated by the fact that they are widely used in wavelet-encoded imaging (\cite{antonini1992image, davis1999wavelet, serrai2005acquisition,liu2012compressive}). 
Therefore, our motivation to study the proposed wavelet-based SRIR is due to the drawbacks of aforementioned methods, growing trend of wavelet-encoded imaging, and the properties of wavelets which include orthogonality, signal localization, and low computing requirements. 

Our approach is very much like single-image super-resolution methods that attempt to extend the spectrum of a single image to higher frequencies, but using information from other images, rather than prior knowledge. In that sense, our method is half-way between single-image and multi-image methods, taking advantage of the best of both worlds. In the scope of this paper, we assume that displacements between the reference and other LR images are known a priori or estimated. We also assume that the displacements are pure translational, or corrected to be so. Our contributions are as follows: \textbf{1.} We establish explicit closed-form expressions that define how the local high frequency information that we aim to recover for the reference HR image is related to the local low frequency information in the given sequence of LR views. \textbf{2.} We assume that the LR images correspond to the polyphase approximation coefficients of the first level wavelet transform of unknown HR images, which allows us to reduce the inverse problem to a collection of well-posed linear problems (local linearity). Our approach is closed-form, and provides results that are superior to the state-of-the-art. We provide the derived formulae utilizing the Haar wavelet transform as an example due to their locality and low computational requirements; however, a general formulation for wavelets can be derived as well. Our exceptional results are attributed to the accuracy, well-posedness, and the linearity of the equations derived in Section \ref{shifts}, and the inherent nature of wavelets, making them very effective in signal localization.

The remainder of this paper is organized as follows. In Section \ref{related}, a brief summary of wavelet-based methods for SRIR is provided. The notations used throughout the paper along with the derived closed-form linear relationships are defined in Section \ref{shifts}. Section \ref{sr} presents the general SRIR observation model, as well as the proposed approach. The stability analysis of the proposed SRIR method is examined in Section \ref{stabil}. Finally, Section \ref{exp} presents the experimental results and comparisons with both single and multi-image state-of-the-art techniques. 

\section{Related Work} \label{related}

Wavelet-based SRIR approaches can be summarized as follows. In order to reduce noise in SRIR methods, \cite{robinson2010efficient} apply a combined Fourier-wavelet deconvolution and denoising algorithm to multiframe SRIR. Authors first produce a sharp and noisy image by fast Fourier based image restoration, then reduce noise by space invariant nonlinear wavelet thresholding. The need to invert large matrices in their method results in solving the problem in the Fourier domain, which has its drawbacks mentioned in Section \ref{intro}. On the other hand, to reduce degradation artifacts such as blurring and the ringing effect, \cite{temizel2005wavelet} utilize zero padding in the wavelet domain followed by cycle spinning. Their method adopts a simplified edge profile and linear regression for edge degradations. Furthermore, to preserve edges, \cite{demirel2011image} utilize stationary and discrete wavelet transforms together in an interpolation-based framework. However, even though better than conventional interpolation techniques, the latter two methods still lack sharp edges.

\cite{zhao2003wavelet} solve a constrained optimization problem utilizing wavelet domain Hidden Markov Tree (HMT) model for the prior knowledge problem, since HMT characterizes the statistics of real world images accurately. In order to suppress the artifacts left after employing their method, cycle spinning is used which leads to blurring as in other interpolation based methods. 

\cite{nguyen2000efficient}, contrary to the conventional interpolation based methods, use the regularity and structure in the interlaced sampling of LR images. Even though, for 2D images, they utilize reshaping property of the Kronecker product, which only doubles the complexity for 1D, their method is based on conjugate gradient which is still time consuming. For deblurring, \cite{chan2003wavelet} derive iterative algorithms, which decompose HR image obtained from an iteration into different frequency components and add them to the next iteration. Their method utilizes wavelet thresholding for denoising, where high-frequency components are penalized, making their method dependent on accurate noise estimation. Moreover, \cite{ji2009robust} handle image registration and reconstruction together, by first estimating the homographies between multiple images, then reconstructing the HR image in a wavelet-based iterative back-projection scheme.

Learning-based methods in wavelet-domain include the following works: \cite{jiji2004single} handle the problem of representing the relationship between LR-HR frames with training their dataset with HR images by learning from wavelet coefficients at finer scales, followed by regularization in a least squares manner; and \cite{gajjar2010new} follow Jiji's method and employ discrete wavelet transform for training, where a cost function based on maximum a posteriori estimation is optimized with gradient descent method, employing an Inhomogeneous Gaussian Markov random field prior. \cite{dong2011image}, to solve the problem of varying contents in different images or image patches, learn various sets of bases from a precollected dataset of example image patches, and select one set of bases adaptively to characterize the local sparse wavelet domain. However, these methods are all based on optimization which requires high computational cost.

The above stated methods are performed either iteratively which requires high computational time or based on interpolation which results in overly smooth images. Our goal is to derive a direct relationship between LR images for a closed-form SRIR solution, which prevents sacrificing high quality. Our paper can be viewed as a generalization of the work by \cite{tong2007super} to some extent. \cite{tong2007super} utilize Taylor series expansion to approximate the high frequency information that we want to recover. Their method is constrained to use LR images which have specific translations (namely 1 pixel in horizontal and 1 pixel in vertical directions). Our method, on the other hand, generalizes the translations for any shift; while determining the exact relationship between LR images and the subbands, instead of providing an approximation. 

\section{Subpixel Shifts of a Low Resolution Image} \label{shifts}

In-band (i.e. wavelet domain) subpixel shift method along with the related notation are provided in this section.

\subsection{Notation} \label{term}

Here, we provide the notations used throughout the paper in Table \ref{termtable}. 

\begin{center}
	\begin{table}[h] 
		\centering
		\caption{Notation}
		\begin{tabular}{l p{0.7\linewidth}}
			$I(2m, 2n)$ & Reference HR image\\
			$A, H, V, D$ & $1st$ level Haar wavelet transform approximation, horizontal, vertical, and diagonal detail coefficients of $I(2m, 2n)$, respectively \\
			$F, K$ & Matrices to be multiplied by approximation and detail coefficients (i.e. $A, H, V, D$) of the reference HR image, that are used to define in-band shift of the reference LR image (i.e. $A$) \\
			$\ell$ & Number of hypothetically added levels in case of non-integer shifts\\
			$s$ & Integer shift amount after the hypothetically added levels ($\ell$)\\
		\end{tabular} \label{termtable}
	\end{table} 
\end{center}

Bold uppercase letters in the following sections demonstrate matrices whereas bold lowercase ones indicate vectors. The subscripts $h, v, d$ demonstrate horizontal, vertical, and diagonal translations, respectively. Finally, the subscript $k$ indicates the $k$th LR or HR image.

\subsection{Subpixel Shifts}

Our goal for the proposed SRIR method is to reconstruct the lost high frequency information of an unknown HR image, given a sequence of subpixel shifted LR images. For this purpose, we first derive the relationship that relates these LR images to the high frequency information of the unknown HR image. This relationship can be described by in-band shift (i.e. in the wavelet domain) of a reference LR image. 

\begin{figure}[t]
	\centering
	\centerline{\includegraphics[width=0.7\textwidth]{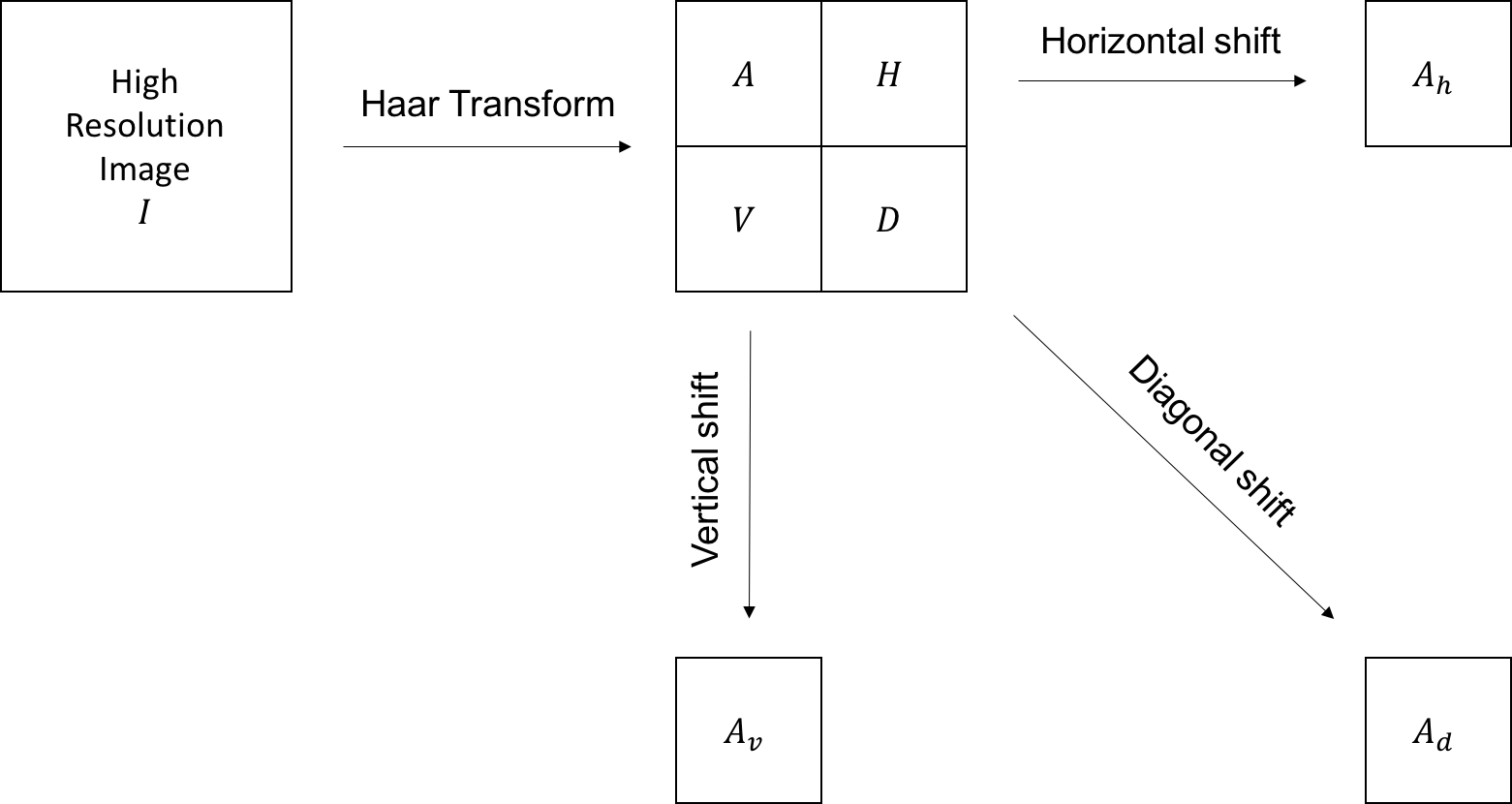}}
	\caption{In-band shift of a reference LR image (i.e. $A$) in the Haar domain, using high frequency information of the related HR image (i.e. $H,V,D$)}\medskip \label{fig:illustrateshift}
\end{figure}

In order to find the aforementioned relationship, we first assume that the reference HR image is known. The reference LR image is the approximation coefficients obtained by decomposing the HR image for 1-level Haar Transform. Then, we define shifted LR images based on the resultant Haar coefficients of the HR image. The shifting process is illustrated in Fig. \ref{fig:illustrateshift}, where shifted LR images (i.e. $A_h, A_v, A_d$) are described based on the first level approximation and detail coefficients of the reference HR image (i.e. $A, H, V, D$).

Below, we derive the mathematical expressions which demonstrate this relationship. The derived equations relate the high-frequency part (i.e. detail wavelet coefficients) of a reference HR image to the low-frequency information provided by the LR image sequence. 

Let $\textbf{A}$, $\textbf{H}$, $\textbf{V}$, and $\textbf{D}$ be the first level approximation (i.e. reference LR image), horizontal, vertical, and diagonal detail coefficients, respectively, of a $2D$ reference HR image, $I(2m, 2n)$, of size $2m\times2n$, where $m$ and $n$ are positive integers. Since 1-level wavelet transform reduces the size of HR image by half in each direction for approximation and detail coefficients, we require the size of HR image to be divisible by 2. Now, a translated LR image in an arbitrary direction can be expressed in matrix form using the $1st$ level Haar transform of $I(2m, 2n)$ as in the following equation in (\ref{firsteq}).

\begin{eqnarray}\label{firsteq} 
\textbf{A}_s &=& \textbf{F}_v \textbf{A} \textbf{F}_h + \textbf{F}_v \textbf{H} \textbf{K}_h + \textbf{K}_v \textbf{V} \textbf{F}_h + \textbf{K}_v \textbf{D} \textbf{K}_h
\end{eqnarray} 

As already mentioned in Section \ref{term}, $\textbf{F}$ and $\textbf{K}$ stand for matrices to be multiplied by the first level lowpass and highpass subbands of the reference HR image, where subscripts $h$ and $v$ indicate \textit{horizontal} and \textit{vertical} shifts. $\textbf{A}_s$ stands for a shifted image in any direction. The low/high-pass subbands together with $\textbf{A}_s$ are of size $m \times n$, $\textbf{F}_v$ and $\textbf{K}_v$ are $m \times m$, whereas $\textbf{F}_h$ and $\textbf{K}_h$ are $n \times n$.

By examining the translational shifts between two LR images in the Haar domain, we realize that horizontal translation reduces $\textbf{K}_v$ to zero and $\textbf{F}_v$ to the identity matrix. This could be comprehended by examining the coefficient matrices defined later in this section (namely, Eq. (\ref{coefmat})), by making related vertical components zero (specifically, $s_y$ and $\ell_y$). This observation lets us define a horizontally shifted image $\textbf{A}_h$ by using only approximation and horizontal detail coefficients. Likewise, vertical translation solely necessitates approximation and vertical detail coefficients, in which case $\textbf{K}_h$ is reduced to zero and $\textbf{F}_h$ is equal to the identity matrix. As a result, the equation shown above in Eq. (\ref{firsteq}), can be expressed for each translation direction as in Eq. (\ref{linear}):

\begin{eqnarray}
\textbf{A}_h &=& \textbf{A} \textbf{F}_h + \textbf{H} \textbf{K}_h \nonumber \\
\textbf{A}_v &=& \textbf{F}_v \textbf{A} + \textbf{K}_v \textbf{V} \nonumber \\ \nonumber
\textbf{A}_d &=& \textbf{F}_v \textbf{A} \textbf{F}_h + \textbf{F}_v \textbf{H} \textbf{K}_h + \textbf{K}_v \textbf{V} \textbf{F}_h + \textbf{K}_v \textbf{D} \textbf{K}_h \\ 
\label{linear}
\end{eqnarray} 

\begin{figure}[h]
	\centering
	\centerline{\includegraphics[width=0.8\textwidth]{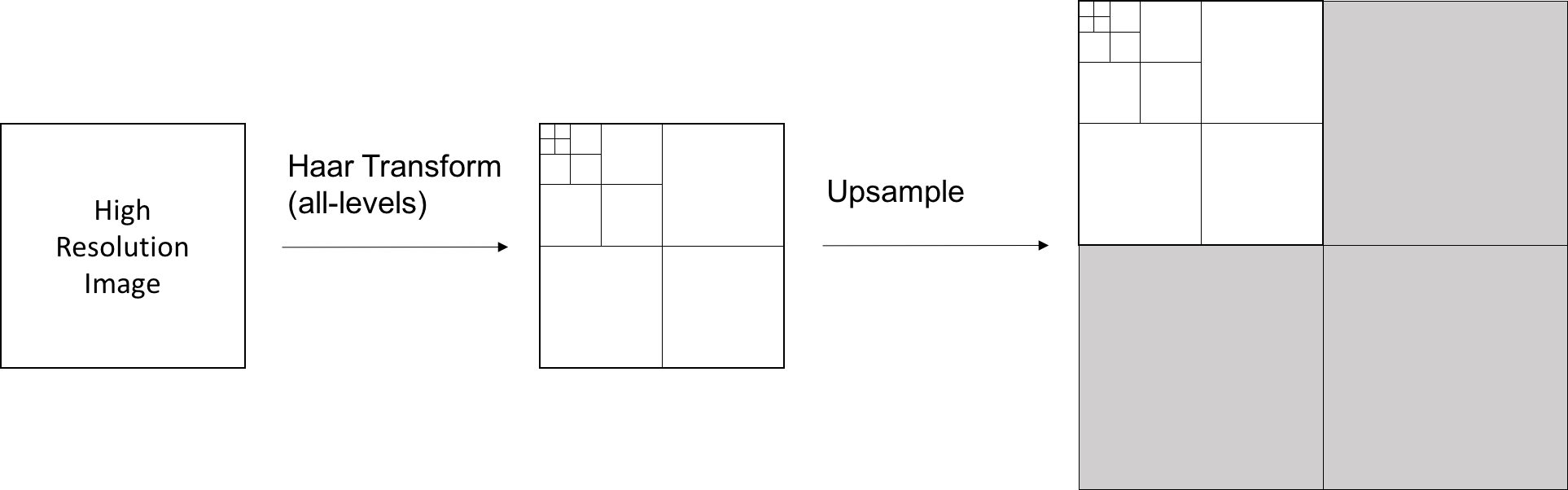}}
	\caption{Upsampling illustration}\medskip \label{fig:upsample}
\end{figure}

Here, our focus is on subpixel translations. Contrary to the general concept of approximating a subpixel shift by upsampling an image followed by an integer shift, our method models subpixel shift directly on the original coefficients of the reference HR image, without upsampling. We observe that: 

\textbf{(1)} Viewing the wavelet transform of the image $I(2m,2n)$ as in Fig. \ref{fig:upsample}, upsampling an image is equivalent to adding levels to the bottom of the transform, and setting the detail coefficients to zero while the approximation coefficients remain the same. 

\textbf{(2)} Shifting the upsampled image by an amount of $s$ is a counterpart of shifting the original image by an amount of $s/2^\ell$, where $\ell$ is the number of added levels. 

Fig. \ref{fig:upsample} demonstrates an example of the upsampling process described above where $\ell=1$, which implies that only 1 level of zero detail coefficients are added. Assuming that the HR image is given, Haar Transform of this HR image can be found readily. For upsampling, these Haar Transform coefficients are utilized as approximation coefficients with more levels of detail coefficients which are set to be zero. Here, gray boxes demonstrate added zeros. 

These observations allow us to shift a reference LR image in-band (when the corresponding HR image is given) for a subpixel amount without actually upsampling it, which saves memory and reduces the computational cost. In order  to shift the reference LR image, the original approximation and detail coefficients of the reference HR image are utilized with a hypothetically added level ($\ell$) and an integer shift value ($s$) at the added level.

Now, the aforementioned matrices, $\textbf{F}_h$, $\textbf{F}_v$, $\textbf{K}_h$, and $\textbf{K}_v$ can be defined, in bidiagonal Toeplitz matrix form as follows.

\begin{eqnarray} 
\resizebox{0.7\linewidth}{!}{%
	$\textbf{F}_h = \dfrac{1}{2^{\ell_x+1}} 
	\begin{bmatrix}
	2^{\ell_x+1} - \abs{s_x} & &   \\
	\abs{s_x} & 2^{\ell_x+1} - \abs{s_x}   \\
	& \abs{s_x} \\
	& & \ddots & \ddots \\
	& \\
	& & & \abs{s_x} & 2^{\ell_x+1} - \abs{s_x} \\ \nonumber
	\end{bmatrix} \nonumber$
}
\end{eqnarray} 

\begin{eqnarray} 
\resizebox{0.7\linewidth}{!}{%
	$\textbf{F}_v = \dfrac{1}{2^{\ell_y+1}} 
	\begin{bmatrix}
	2^{\ell_y+1} - \abs{s_y} & \abs{s_y}   \\
	& 2^{\ell_y+1} - \abs{s_y} & & \abs{s_y}   \\
	& &  \ddots & \ddots\\
	& \\
	& & & & \abs{s_y} \\
	& & & & 2^{\ell_y+1} - \abs{s_y} \\ \nonumber
	\end{bmatrix} \nonumber$
}
\end{eqnarray} 

\begin{eqnarray}
\textbf{K}_h = \dfrac{1}{2^{\ell_x+1}} 
\begin{bmatrix}
-s_x &  \\
s_x & -s_x   \\
& s_x & \\
& & \ddots & \ddots\\
&\\
& & & s_x & -s_x \\ 
\end{bmatrix} \nonumber
\end{eqnarray}

\begin{eqnarray}
\textbf{K}_v = \dfrac{1}{2^{\ell_y+1}} 
\begin{bmatrix}
-s_y & s_y  & & & \\
& -s_y & s_y   \\
& & \ddots & \ddots\\
& & &  -s_y &  s_y \\
& & & & -s_y \\ 
\end{bmatrix} \label{coefmat}
\end{eqnarray} 

\noindent where $s_{x,y}$ and $\ell_{x,y}$ demonstrate the integer shift amounts at the hypothetically added level and the number of added levels for $x$ and $y$ directions, respectively. 

As mentioned earlier, $\textbf{F}_h$ and $\textbf{K}_h$ are $n \times n$, while $\textbf{F}_v$ and $\textbf{K}_v$ are $m \times m$. Sizes of these matrices also indicate that in-band shift of a reference LR image is performed using only the original level Haar coefficients (which are of size $m \times n$) without upsampling. These matrices show that a 2-pixel neighborhood in the approximation and detail coefficients of a reference HR image is utilized to shift a reference LR image in-band. When the shift amount is negative, diagonals of the matrices interchange. We leave these matrices as square for them to be nonsingular in the SRIR process, otherwise these matrices could be adapted for periodic boundary condition by making them rectangular.

When the shift amount is not divisible by $2^\ell$, in order to reach an integer value at the $(N+\ell)$th level, the shift value at the original level is rounded to the closest decimal point which is divisible by $2^\ell$. 

Here, derived matrices to calculate the first level lowpass subband of a shifted image (i.e. $\textbf{A}_h, \textbf{A}_v, \textbf{A}_d$) are demonstrated. The counterparts for the first level detail coefficients (e.g. $\textbf{H}_h, \textbf{H}_v, \textbf{H}_d, \textbf{V}_h,...$) of the shifted HR images can be found in a similar manner, in order to shift the entire HR reference image directly in the Haar domain. Since for our SRIR method, we will only employ the relationship for approximation coefficients (i.e. LR images), we only provide the related equations.

\section{Super Resolution Image Reconstruction} \label{sr}

In this section, we first present the SRIR observation model, followed by our proposed method.

\subsection{Observation Model} \label{obs}

Let $I(2m,2n)$ denote the desired HR image, and $A_k$ be the $k$th observed LR image. The super resolution observation model is given by:

\begin{eqnarray}
\textbf{a}_k = \mathbf{\Lambda}_k \mathbf{B}_k \textbf{M}_k \textbf{i} + \textbf{n}_k,\quad k = 1,2,...,K
\end{eqnarray}

\noindent where $\textbf{M}_k$, $\textbf{B}_k$, $\mathbf{\Lambda}_k$, and $\textbf{n}_k$ denote motion, blurring effect, downsampling operator, and noise term for the $k$th LR image, respectively, and $K$ is the number of observed LR images. $\textbf{a}_k$ and $\textbf{i}$ are the $k$th LR image and unknown HR images, respectively, represented in lexicographical order.

Given a sequence of observed LR images, $\textbf{a}_k$, the goal of SRIR is to reconstruct an unknown HR image, $\textbf{i}$.

\subsection{Proposed Method}

As in the underlying idea of wavelet-based SRIR algorithms, we assume that the given LR image sequence is the lowpass subbands (i.e. approximation coefficients of 1-level Wavelet Transform) of unknown HR images. The goal is to reconstruct the unknown highpass subbands (i.e. detail coefficients of 1-level Wavelet Transform) of one of these HR images which is chosen as the reference one. The SRIR method described below is the inverse process of the method described in Section \ref{shifts}, where HR images are unknown, and high frequency information for one of these underlying HR images is estimated by solving a related linear system.

The relationship between two subpixel shifted LR images depends on the highpass subbands of the underlying reference HR image, as demonstrated in the previous section. This fact is used to construct a linear system of equations based on known LR images (i.e. $\textbf{A}, \textbf{A}_h, \textbf{A}_v, \textbf{A}_d$ in Section \ref{shifts}) and unknown highpass subbands of the reference HR image (i.e. $\textbf{H},\textbf{V},\textbf{D}$ in Section \ref{shifts}) using related formulae from Eq. (\ref{linear}) depending on the translation direction. Since there are three unknowns (i.e. horizontal, vertical, and diagonal detail coefficients of the unknown HR image), three shifted LR images together with the reference LR image are required to solve the linear system. Once this system is solved for the unknowns, inverse Haar transform utilizing the reference LR image and the estimated highpass subbands of the underlying unknown reference HR image gives the reconstructed HR image. 

\begin{figure}[h]
	\centering
	\centerline{\includegraphics[width=\textwidth]{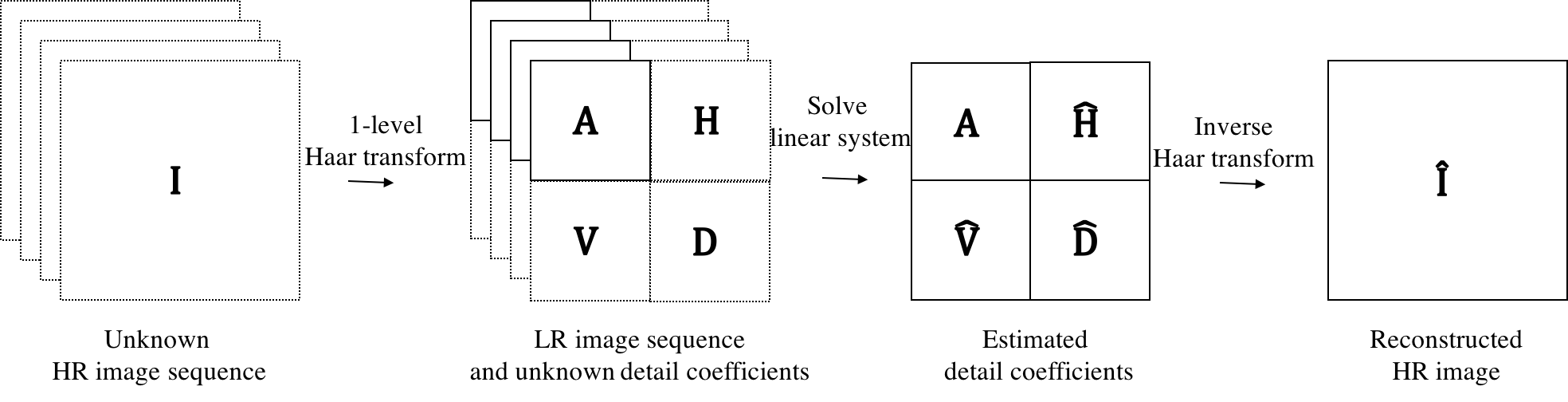}}
	\caption{Proposed method for Super Resolution Image Reconstruction}\medskip \label{fig:srprocess}
\end{figure}

Fig. \ref{fig:srprocess} shows a pictorial explanation of the proposed method, where solid boxes indicate known or estimated coefficients and dotted boxes show unknown ones. Images with the hat symbol (i.e. $\hat{}$ ) stands for estimated coefficients. As the figure demonstrates, assuming the LR sequence is first level approximation coefficients of the wavelet transform, we estimate the unknown high frequency information of the reference HR image in order to reconstruct the estimated HR image. 

In the scope of this paper, we assume that the registration between images are known a priori or has been estimated. Translational shifts can be estimated using one of the methods by \cite{foroosh2002extension, shekarforoush1996subpixel, foroosh2004sub, balci2005estimating, balci2005inferring, balci2006subpixel, balci2006subpixel2, evangelidis2008parametric, vandewalle2006frequency}. Even though the equations derived in Section \ref{shifts} are for subpixel shifts, we apply the proposed SRIR method to the intersection area of any given shift, which may include an integer part, as well. 

The proposed algorithm can also be explained step by step in \textbf{Algorithm} - Super Resolution Image Reconstruction as follows.

\begin{center}
	\begin{table}[h]
		\textbf{Algorithm} \textit{Super Resolution Image Reconstruction}
		\begin{itemize}
			\item \textit{Input}: Observed LR images, registration information 
			\item \textit{Objective}: Estimate high frequency information of the unknown reference HR
			\item \textit{Output}: Reconstructed HR image
		\end{itemize}
		\begin{itemize} 
			\item[$\blacktriangleright$] Imaging Process 
			\begin{itemize}
				\item[$\diamond$] Shift a given High Resolution image randomly in order to obtain three more HR images
				\item[$\diamond$] Transform all four HR images in the Haar domain (1-level) to acquire observed shifted LR images
				\item[$\diamond$] If shift amount is not subpixel, remove integer parts to find intersection of images which can be defined as subpixel translation
			\end{itemize}  
			\item[$\blacktriangleright$] Super Resolution Image Reconstruction Process
			\begin{itemize}
				\item[$\diamond$] Solve a linear system of equations comprised of the formulae in Eq. (\ref{linear}) based on the direction of the translation, for highpass subbands $H$, $V$, and $D$ of the reference HR image, using observed LR images and known displacements
				\item[$\diamond$] Perform inverse Haar Transform on the reference LR image and estimated highpass subbands (i.e. detail coefficients) to reconstruct the HR image
			\end{itemize}
		\end{itemize}
	\end{table}
\end{center}

\section{Stability Analysis} \label{stabil}

In this section, we will investigate the stability of our method.

As mentioned in Section \ref{sr}, our method constructs a linear system of equations based on given LR images and related shifts. Since the LR images (i.e. $\textbf{A}$, $\textbf{A}_h$, $\textbf{A}_v$, and $\textbf{A}_d$) and the displacements between them are known, $\textbf{H}, \textbf{V}$, and $\textbf{D}$ are the only unknowns of the constructed system. This linear system may appear in four forms which include:

\begin{itemize}
	\item[$\circ$] 1 horizontally, 1 vertically, 1 diagonally shifted image
	\item[$\circ$] 1 horizontally, 2 diagonally shifted images
	\item[$\circ$] 1 vertically, 2 diagonally shifted images
	\item[$\circ$] 3 diagonally shifted images
\end{itemize}
\noindent along with the reference LR image, $\textbf{A}$, where second and third cases demonstrate the same properties. Thus, we will consider the first, second, and last cases in our analysis.

\textbf{Case 1} \textit{(1 horizontal, 1 vertical, 1 diagonal)} \textbf{:} This case constructs a linear system of equations exactly as shown in Eq. (\ref{linear}). This linear system is solved first for $\textbf{H}$ using the equation for $\textbf{A}_h$, then for $\textbf{V}$ using the equation for $\textbf{A}_v$, and finally for $\textbf{D}$ using the equation for $\textbf{A}_d$ and substituting the information found for $\textbf{H}$ and $\textbf{V}$. Since the coefficient matrices are invertible, this system is stable.

\textbf{Case 2 and 3} \textit{(1 horizontal/vertical, 2 diagonal)} \textbf{:} Here, we will explore Case 2 with 1 horizontally and 2 diagonally shifted images. Case 3 will demonstrate similar features as mentioned above.

This case includes one $\textbf{A}_h$ and two $\textbf{A}_d$ from Eq. (\ref{linear}) for one horizontal and two diagonally shifted images, where the linear system takes the form:

\begin{eqnarray}
\textbf{A}_h &=& \textbf{A} \textbf{F}_h + \textbf{H} \textbf{K}_h \nonumber \\ \nonumber
\textbf{A}_{d_1} &=& \textbf{F}_{v_1} \textbf{A} \textbf{F}_{h_1} + \textbf{F}_{v_1} \textbf{H} \textbf{K}_{h_1} + \textbf{K}_{v_1} \textbf{V} \textbf{F}_{h_1} + \textbf{K}_{v_1} \textbf{D} \textbf{K}_{h_1} \\ \nonumber
\textbf{A}_{d_2} &=& \textbf{F}_{v_2} \textbf{A} \textbf{F}_{h_2} + \textbf{F}_{v_2} \textbf{H} \textbf{K}_{h_2} + \textbf{K}_{v_2} \textbf{V} \textbf{F}_{h_2} + \textbf{K}_{v_2} \textbf{D} \textbf{K}_{h_2}  \\ 
\end{eqnarray} 

Again, as in Case 1, the first equation is stable, therefore $\textbf{H}$ can be found easily. Solving equations for $\textbf{A}_{d_1}$ and $\textbf{A}_{d_2}$ for $\textbf{V}$ results in:

\begin{eqnarray}\label{b}
\textbf{V} (\textbf{F}_{h_1} {\textbf{K}_{h_1}}^{-1} - \textbf{F}_{h_2} {\textbf{K}_{h_2}}^{-1}) =
{\textbf{K}_{v_1}}^{-1} \textbf{S} {\textbf{K}_{h_1}}^{-1}
- {\textbf{K}_{v_2}}^{-1} \textbf{T} {\textbf{K}_{h_2}}^{-1}
\end{eqnarray} 

\noindent where 

\begin{eqnarray}
\textbf{S} = \textbf{A}_{d_1} - \textbf{F}_{v_1} \textbf{A} \textbf{F}_{h_1} - \textbf{F}_{v_1} \textbf{H} \textbf{K}_{h_1}\nonumber \\ 
\textbf{T} = \textbf{A}_{d_2} - \textbf{F}_{v_2} \textbf{A} \textbf{F}_{h_2} - \textbf{F}_{v_2} \textbf{H} \textbf{K}_{h_2} \nonumber\\\nonumber
\end{eqnarray} 

In order to tackle the instability problem caused by inverting multiplication and summation of matrices in Eq. (\ref{b}), we right multiply this equation with $\textbf{K}_{h_1}$. Since $\textbf{K}_{h_1}$ and $\textbf{K}_{h_2}$ differ only by the shift value for the two diagonally shifted images, ${\textbf{K}_{h_2}}^{-1} \textbf{K}_{h_1}$ results in the identity matrix multiplied by a scalar which depends only on the shifts. Thus, the equation for $\textbf{V}$ becomes:

\begin{eqnarray}\label{bfinal}
\textbf{V} (\textbf{F}_{h_1} - \textbf{F}_{h_2} \alpha) =
{\textbf{K}_{v_1}}^{-1} \textbf{S}
- {\textbf{K}_{v_2}}^{-1} \textbf{T} \alpha
\end{eqnarray} 

\noindent where

\begin{eqnarray}
\alpha = {\textbf{K}_{h_2}}^{-1} {\textbf{K}_{h_1}} \nonumber
\end{eqnarray} 

\noindent where $\alpha$ is defined as a constant.

Truncated Singular Value Decomposition (TSVD) is used with the resulting equation in (\ref{bfinal}) to find $\textbf{V}$. Rank of TSVD method is decided based on minimizing the following cost function:

\begin{eqnarray} \label{cost}
\operatorname*{arg\,min}_{k} ||\textbf{X}^{-1}_r \textbf{X}_r - \textbf{U}||_F
\end{eqnarray}

\noindent where $\textbf{U}$ shows the identity (i.e. unit) matrix, subscript $F$ is \textit{Frobenius norm}, and $\textbf{X}_r$ stands for \textit{rank-r} approximation of a matrix $\textbf{X}$.

In order to successfully truncate $\textbf{X}$ at $r$, we follow a theorem by \cite{hansen1987truncatedsvd} (Theorem 3.2), which implies that there must be a well-determined gap between the two consecutive singular values at $r$ (i.e. $\sigma_r$) and $r+1$ (i.e. $\sigma_{r+1}$).

As one can see in Eq. (\ref{bfinal}), the stability of our method is partially dependent on the closeness of shift amounts. 

\textbf{Case 4} \textit{(3 diagonal)} \textbf{:} The final case includes three diagonally shifted images together with the reference image. Therefore, the linear system is constructed as:

\begin{eqnarray}\label{3d}
\textbf{A}_{d_1} &=& \textbf{F}_{v_1} \textbf{A} \textbf{F}_{h_1} + \textbf{F}_{v_1} \textbf{H} \textbf{K}_{h_1} + \textbf{K}_{v_1} \textbf{V} \textbf{F}_{h_1} + \textbf{K}_{v_1} \textbf{D} \textbf{K}_{h_1} \nonumber\\ 
\textbf{A}_{d_2} &=& \textbf{F}_{v_2} \textbf{A} \textbf{F}_{h_2} + \textbf{F}_{v_2} \textbf{H} \textbf{K}_{h_2} + \textbf{K}_{v_2} \textbf{V} \textbf{F}_{h_2} + \textbf{K}_{v_2} \textbf{D} \textbf{K}_{h_2}  \nonumber\\ 
\textbf{A}_{d_3} &=& \textbf{F}_{v_3} \textbf{A} \textbf{F}_{h_3} + \textbf{F}_{v_3} \textbf{H} \textbf{K}_{h_3} + \textbf{K}_{v_3} \textbf{V} \textbf{F}_{h_3} + \textbf{K}_{v_3} \textbf{D} \textbf{K}_{h_3} \nonumber\\ 
\end{eqnarray} 

By solving the system above in Eq. (\ref{3d}) for $\textbf{H}$, we find a generalized Sylvester equation as in:

\begin{eqnarray}\label{sylvester}
\textbf{P}_1 \textbf{H} \textbf{Q}_1 - \textbf{P}_2 \textbf{H}  \textbf{Q}_2 = \textbf{R} 
\end{eqnarray}

\noindent where

\begin{eqnarray}
\textbf{P}_1 &=& {\textbf{K}_{v_1}}^{-1} (\textbf{F}_{v_1} - \textbf{K}_{v_1} {\textbf{K}_{v_2}}^{-1} \textbf{F}_{v_2}) \nonumber\\
\textbf{P}_2 &=& {\textbf{K}_{v_2}}^{-1} (\textbf{F}_{v_2} - \textbf{K}_{v_2} {\textbf{K}_{v_3}}^{-1} \textbf{F}_{v_3})\nonumber\\
\textbf{Q}_1 &=& \textbf{K}_{h_1} (\textbf{F}_{h_1} - \textbf{F}_{h_2} {\textbf{K}_{h_2}}^{-1} \textbf{K}_{h_1})^{-1} \nonumber\\
\textbf{Q}_2 &=& \textbf{K}_{h_2} (\textbf{F}_{h_2} - \textbf{F}_{h_3} {\textbf{K}_{h_3}}^{-1} \textbf{K}_{h_2})^{-1}  \nonumber\\
\textbf{R} &=& [({\textbf{K}_{v_1}}^{-1} \textbf{A}_{d_1} - {\textbf{K}_{v_1}}^{-1} \textbf{F}_{v_1} \textbf{A} \textbf{F}_{h_1})  \nonumber\\ &&-  ({\textbf{K}_{v_2}}^{-1} \textbf{A}_{d_2} - {\textbf{K}_{v_2}}^{-1} \textbf{F}_{v_2} \textbf{A} \textbf{F}_{h_2}) {\textbf{K}_{h_2}}^{-1} \textbf{K}_{h_1}]  \nonumber\\
&\times& (\textbf{F}_{h_1} - \textbf{F}_{h_2} {\textbf{K}_{h_2}}^{-1} \textbf{K}_{h_1})^{-1}  \nonumber\\
&-& [({\textbf{K}_{v_2}}^{-1} \textbf{A}_{d_2} - {\textbf{K}_{v_2}}^{-1} \textbf{F}_{v_2} \textbf{A} \textbf{F}_{h_2})  \nonumber\\ &&-  ({\textbf{K}_{v_3}}^{-1} \textbf{A}_{d_3} - {\textbf{K}_{v_3}}^{-1} \textbf{F}_{v_3} \textbf{A} \textbf{F}_{h_3}) {\textbf{K}_{h_3}}^{-1} \textbf{K}_{h_2}]  \nonumber\\
&\times& (\textbf{F}_{h_2} - \textbf{F}_{h_3} {\textbf{K}_{h_3}}^{-1} \textbf{K}_{h_2})^{-1} \nonumber
\end{eqnarray} 

By examining $\textbf{P}_i$ for $i = 1,2$, in the generalized Sylvester equation, $\textbf{K}_{v_i} {\textbf{K}_{v_{i+1}}}^{-1}$ could be changed by multiplication by a scalar (as in ${\textbf{K}_{h_2}}^{-1} {\textbf{K}_{h_1}}$ in Eq. (\ref{bfinal})), which leaves $\textbf{F}_{v_i} - \textbf{K}_{v_i} {\textbf{K}_{v_{i+1}}}^{-1} \textbf{F}_{v_{i+1}}$ as an upper bidiagonal matrix, since $\textbf{F}_{v_i}$ is also upper bidiagonal. Moreover, since $\textbf{K}_v$ is an upper bidiagonal matrix, inverse of $\textbf{K}_v$ is an upper triangular matrix (\cite{vandebril2007matrix}). Therefore, by multiplication of two upper triangular matrices, we obtain upper triangular matrices for $\textbf{P}_i$. By following similar analysis, we observe that $\textbf{Q}_i$ are lower triangular matrices.

Here, we refer to a theorem by \cite{chu1987solution} for a generalized Sylvester equation to have a unique solution. Interested reader can find the proof for this theorem in the referred paper; we include the theorem here to make this paper self-contained.

\textbf{Theorem:} The matrix equation in (\ref{sylvester}) has a unique solution if and only if 
\begin{enumerate}
	\item $\textbf{P}_1 - \lambda \textbf{P}_2$ and $\textbf{Q}_2 - \lambda \textbf{Q}_1$ are regular matrix pencils, \\
	and
	\item $\rho(\textbf{P}_1, \textbf{P}_2) \cap \rho(\textbf{Q}_1, \textbf{Q}_2) = \emptyset$
\end{enumerate}

\noindent where $\lambda$ shows the generalized eigenvalues of the matrix pencils, $\rho$ defines the spectra of the generalized eigenvalues, and $(.,.)$ demonstrates a matrix pencil.

The matrix pencils constructed as $(\textbf{P}_1, \textbf{P}_2)$, and $(\textbf{Q}_1, \textbf{Q}_2)$, using the given $\textbf{P}_1, \textbf{P}_2, \textbf{Q}_1$ and $\textbf{Q}_2$ in Eq. (\ref{sylvester}), are not guaranteed either to be regular, or to have empty intersection of generalized eigenvalue spectra. For instance, when any of the two LR images have negative horizontal shift amount, the related $\textbf{Q}_i$ has zero diagonals, and a zero element on the diagonal makes the matrix pencil singular when a matrix pencil is upper/lower triangular (\cite{bai2000templates}). Since we know that $\textbf{P}_i$, and $\textbf{Q}_i$, for $i=1,2$, are upper and lower triangular matrices, respectively, forming upper/lower triangular matrix pencils, two images with negative horizontal shifts satisfy requirements for singular matrix pencils. 

Based on these facts, solution methods utilized for generalized Sylvester equation cannot be employed here. Therefore, in order to find a solution to the system in Eq. (\ref{3d}), we first vectorize the equations using Kronecker tensor product, before solving for the unknowns:

\begin{eqnarray} \label{kron}
\textbf{a}_{d_i} &=& (\textbf{F}_{v_i} \otimes \textbf{F}_{h_i}') \textbf{a} + (\textbf{F}_{v_i} \otimes \textbf{K}_{h_i}') \textbf{h} + (\textbf{K}_{v_i} \otimes \textbf{F}_{h_i}') \textbf{v} + (\textbf{K}_{v_i} \otimes \textbf{K}_{h_i}') \textbf{d} \nonumber\\
\end{eqnarray}

\noindent for $i = 1,2,3$. Here, lowercase bold letters indicate column-vise vectorized versions of $\textbf{A}_{d_i}, \textbf{A}, \textbf{H}, \textbf{V}$, and $\textbf{D}$, and these vectors have size $mn \times 1$. The Kronecker tensor products in parenthesis result in matrices of size $mn \times mn$, where $m$ and $n$ are the size of LR images.

By solving Eq. (\ref{kron}) for $\textbf{h}$, we find the following equation which appears similar to the equation for \textbf{V} in Eq. (\ref{b}):

\begin{eqnarray} \label{a}
[\textbf{W}_1^{-1} \textbf{Y}_1 - \textbf{W}_2^{-1} \textbf{Y}_2] \textbf{h} = \textbf{W}_1^{-1} \textbf{Z}_1 - \textbf{W}_2^{-1} \textbf{Z}_2
\end{eqnarray}

\noindent where 
\begin{eqnarray} 
\textbf{W}_i &=& (\textbf{KF})_i - (\textbf{KK})_i {(\textbf{KK})^{-1}_{i+1}} (\textbf{KF})_{i+1}  \nonumber\\
\textbf{Y}_i &=& (\textbf{FK})_i - (\textbf{KK})_i {(\textbf{KK})^{-1}_{i+1}} (\textbf{FK})_{i+1} \nonumber\\
\textbf{Z}_i &=& \textbf{a}_{d_i} - (\textbf{KK})_i {(\textbf{KK})^{-1}_{i+1}} \textbf{a}_{d_{i+1}} \nonumber\\&&- [(\textbf{FF})_i - (\textbf{KK})_i {(\textbf{KK})^{-1}_{i+1}} (\textbf{FF})_{i+1}] \textbf{a}  \nonumber
\end{eqnarray}
\noindent for $i=1,2$, and
\begin{eqnarray}
\textbf{FF} &=& \textbf{F}_v \otimes \textbf{F}_h' \nonumber \\
\textbf{FK} &=& \textbf{F}_v \otimes \textbf{K}_h' \nonumber\\
\textbf{KF} &=& \textbf{K}_v \otimes \textbf{F}_h' \nonumber\\ 
\textbf{KK} &=& \textbf{K}_v \otimes \textbf{K}_h' \nonumber
\end{eqnarray} 

Here, in order to solve for $\textbf{h}$, we follow a similar approach to the one used to reach Eq. (\ref{bfinal}) from Eq. (\ref{b}), where Eq. (\ref{a}) is left multiplied by $\textbf{W}_1$ in order to reduce the instability. Again, TSVD is utilized to solve the equation with the same cost function used in Eq. (\ref{cost}). 

As in Cases 2 and 3, the stability of our solution depends partly on the closeness of shift values which affects the matrix inversions.

\section{Experimental Results} \label{exp}

In this section, we first present the implementation details, followed by results for the proposed method along with comparisons to the recent state-of-the-art and conventional techniques. Comparisons are made based on qualitative and quantitative evaluations on both commonly adapted test examples and real world images to demonstrate the influence of compression artifacts and sensor noise on the proposed method. LR image sequences are synthetically generated. Computational time efficiency of the proposed method against other methods are also presented. Moreover, HR and LR reference images for all test cases and zoomed parts in detailed areas for each image are provided. 

\subsection{Implementation Details}

LR image sequences are synthesized by the method explained in Section \ref{sr} (Imaging Process in \textbf{Algorithm}). LR images are divided into overlapping blocks of size $32 \times 32$, in order to reduce memory usage and decrease computational time. 

To simulate the motion estimation error for the proposed method, HR reference image is shifted randomly for a shift amount which is not necessarily divisible by $2^\ell$ and shifts are rounded to the closest decimal divisible by $2^\ell$ for the calculations, as described before in Section \ref{shifts}.

For the cases when the shift amounts are not subpixel (which might be integer or include an integer part), we find the intersection area of the images which can be described as subpixel shift. We apply the same method to the intersected area, where boundaries are lost for the maximum integer amount among all shifts.

In order to reduce the boundary problem caused by square coefficient matrices $\textbf{F}_{h,v}$ and $\textbf{K}_{h,v}$ which does not include the information in the boundaries, the last rows and columns of calculated $\textbf{H}, \textbf{V}$, and $\textbf{D}$ are extrapolated.

Color images are handled by the conventional approach (\cite{dong2011image,yang2010image}), in \textit{YIQ} color space, where only the illuminance channel of images are dealt with the proposed method, since human visual system is more sensitive to changes in illuminance channel. The chrominance channels are upsampled using bicubic interpolation. 

\subsection{Qualitative Comparison} 

We compared our method with both multiframe and single image SRIR techniques including interpolation-based ones which are Bicubic interpolation and Robust Super Resolution (\cite{zomet2001robust}), a regularization-based methods by \cite{babacan2011variational} and \cite{farsiu2004fast}, and finally a wavelet-domain learning-based method by \cite{dong2011image}. Compared methods were given the same input images and knowledge of registration (if required).

\begin{figure*}[t]
	\begin{minipage}[b]{0.2\linewidth}
		\centering
		\centerline{\includegraphics[width=3.2cm]{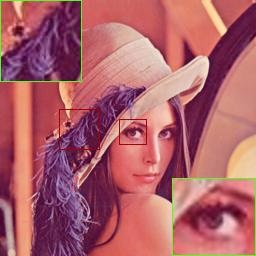}}
		\centerline{\textbf{(a)} Lena HR image}\medskip
	\end{minipage}
	\hfill
	\begin{minipage}[b]{0.2\linewidth}
		\centering
		\centerline{\includegraphics[width=3.2cm]{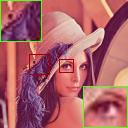}}
		\centerline{\textbf{(b)} LR reference image}\medskip
	\end{minipage}
	\hfill
	\begin{minipage}[b]{0.2\linewidth}
		\centering
		\centerline{\includegraphics[width=3.2cm]{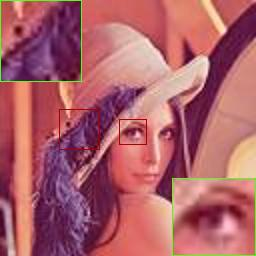}}
		\centerline{\textbf{(c)} Bicubic}\medskip
	\end{minipage}
	\hfill
	\begin{minipage}[b]{0.2\linewidth}
		\centering
		\centerline{\includegraphics[width=3.2cm]{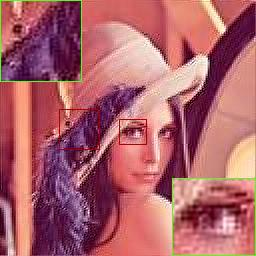}}
		\centerline{\textbf{(d)} \cite{zomet2001robust}}\medskip
	\end{minipage}
	\hfill
	\begin{minipage}[b]{0.2\linewidth}
		\centering
		\centerline{\includegraphics[width=3.2cm]{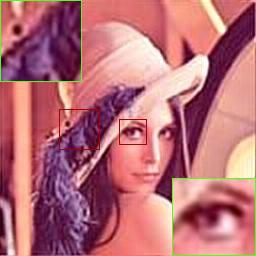}}
		\centerline{\textbf{(e)} \cite{babacan2011variational}}\medskip
	\end{minipage}
	\hfill
	\begin{minipage}[b]{0.2\linewidth}
		\centering
		\centerline{\includegraphics[width=3.2cm]{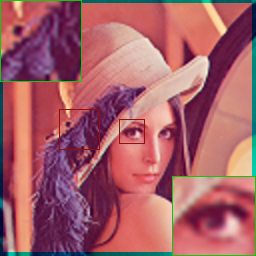}}
		\centerline{\textbf{(f)} \cite{farsiu2004fast}}\medskip
	\end{minipage}
	\hfill
	\begin{minipage}[b]{0.2\linewidth}
		\centering
		\centerline{\includegraphics[width=3.2cm]{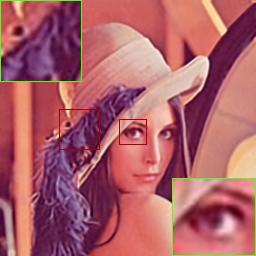}}
		\centerline{\textbf{(g)} \cite{dong2011image}}\medskip
	\end{minipage}
	\hfill
	\begin{minipage}[b]{0.2\linewidth}
		\centering
		\centerline{\includegraphics[width=3.2cm]{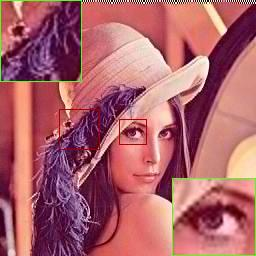}}
		\centerline{\textbf{(h)} Ours}\medskip
	\end{minipage}
	\caption{Lena image comparison ($\times 2$), with zoomed parts in green.}
	\label{fig:lena}
\end{figure*}

\begin{figure*}[t]
	\begin{minipage}[b]{0.2\linewidth}
		\centering
		\centerline{\includegraphics[width=3.2cm]{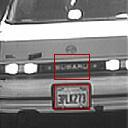}}
		\vspace{0.1cm}
		\includegraphics[width=1.1cm]{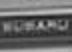}
		\includegraphics[width=1.1cm]{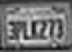}
		\centerline{\textbf{(a)} Car tag HR image}\medskip
	\end{minipage}
	\hfill
	\begin{minipage}[b]{0.2\linewidth}
		\centering
		\centerline{\includegraphics[width=3.2cm]{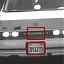}}
		\vspace{0.1cm}
		\includegraphics[width=1.1cm]{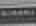}
		\includegraphics[width=1.1cm]{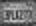}
		\centerline{\textbf{(b)} LR reference image}\medskip
	\end{minipage}
	\hfill
	\begin{minipage}[b]{0.2\linewidth}
		\centering
		\centerline{\includegraphics[width=3.2cm]{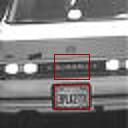}}
		\vspace{0.1cm}
		\includegraphics[width=1.1cm]{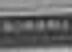}
		\includegraphics[width=1.1cm]{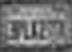}
		\centerline{\textbf{(c)} Bicubic}\medskip
	\end{minipage}
	\hfill
	\begin{minipage}[b]{0.2\linewidth}
		\centering
		\centerline{\includegraphics[width=3.2cm]{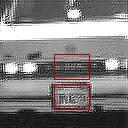}}
		\vspace{0.1cm}
		\includegraphics[width=1.1cm]{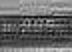}
		\includegraphics[width=1.1cm]{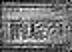}
		\centerline{\textbf{(d)} \cite{zomet2001robust}}\medskip
	\end{minipage}
	\hfill
	\begin{minipage}[b]{0.2\linewidth}
		\centering
		\centerline{\includegraphics[width=3.2cm]{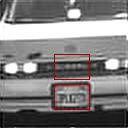}}
		\vspace{0.1cm}
		\includegraphics[width=1.1cm]{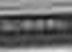}
		\includegraphics[width=1.1cm]{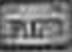}
		\centerline{\textbf{(e)} \cite{babacan2011variational}}\medskip
	\end{minipage}
	\hfill
	\begin{minipage}[b]{0.2\linewidth}
		\centering
		\centerline{\includegraphics[width=3.2cm]{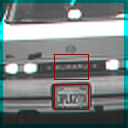}}
		\vspace{0.1cm}
		\includegraphics[width=1.1cm]{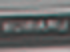}
		\includegraphics[width=1.1cm]{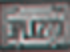}
		\centerline{\textbf{(f)} \cite{farsiu2004fast}}\medskip
	\end{minipage}
	\hfill
	\begin{minipage}[b]{0.2\linewidth}
		\centering
		\centerline{\includegraphics[width=3.2cm]{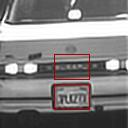}}
		\vspace{0.1cm}
		\includegraphics[width=1.1cm]{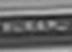}
		\includegraphics[width=1.1cm]{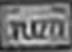}
		\centerline{\textbf{(g)} \cite{dong2011image}}\medskip
	\end{minipage}
	\hfill
	\begin{minipage}[b]{0.2\linewidth}
		\centering
		\centerline{\includegraphics[width=3.2cm]{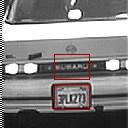}}
		\vspace{0.1cm}
		\includegraphics[width=1.1cm]{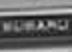}
		\includegraphics[width=1.1cm]{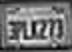}
		\centerline{\textbf{(h)} Ours}\medskip
	\end{minipage}
	\caption{Car tag image (MDSP dataset \cite{farsiu2006multiframe}) comparison ($\times 2$), with zoomed parts below reconstructed ones.}
	\label{fig:tag}
\end{figure*}

\begin{figure*}[h]
	\begin{minipage}[b]{0.2\linewidth}
		\centering
		\centerline{\includegraphics[width=3.2cm]{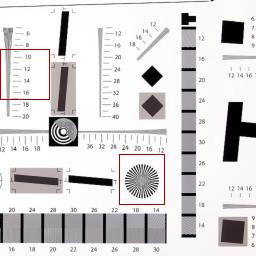}}
		\vspace{0.1cm}
		\includegraphics[width=1.1cm]{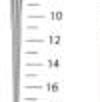}
		\includegraphics[width=1.1cm]{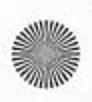}
		\centerline{\textbf{(a)} HR image}\medskip
	\end{minipage}
	\hfill
	\begin{minipage}[b]{0.2\linewidth}
		\centering
		\centerline{\includegraphics[width=3cm]{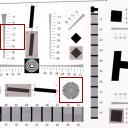}}
		\vspace{0.1cm}
		\includegraphics[width=1.1cm]{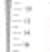}
		\includegraphics[width=1.1cm]{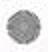}
		\centerline{\textbf{(b)} LR reference image}\medskip
	\end{minipage}
	\hfill
	\begin{minipage}[b]{0.2\linewidth}
		\centering
		\centerline{\includegraphics[width=3cm]{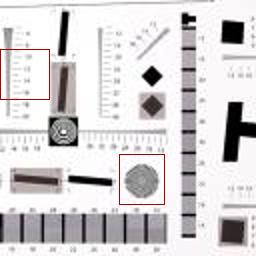}}
		\vspace{0.1cm}
		\includegraphics[width=1.1cm]{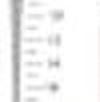}
		\includegraphics[width=1.1cm]{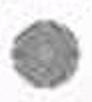}
		\centerline{\textbf{(c)} Bicubic}\medskip
	\end{minipage}
	\hfill
	\begin{minipage}[b]{0.2\linewidth}
		\centering
		\centerline{\includegraphics[width=3cm]{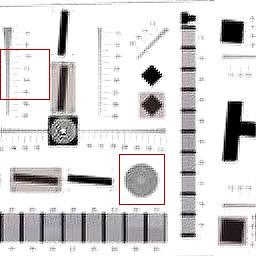}}
		\vspace{0.1cm}
		\includegraphics[width=1.1cm]{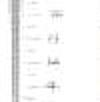}
		\includegraphics[width=1.1cm]{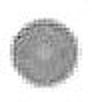}
		\centerline{\textbf{(d)} \cite{zomet2001robust}}\medskip
	\end{minipage}
	\hfill
	\begin{minipage}[b]{0.2\linewidth}
		\centering
		\centerline{\includegraphics[width=3cm]{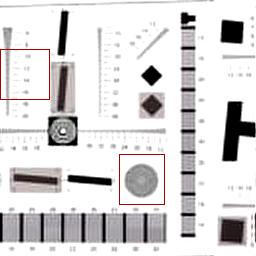}}
		\vspace{0.1cm}
		\includegraphics[width=1.1cm]{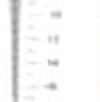}
		\includegraphics[width=1.1cm]{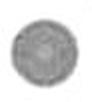}
		\centerline{\textbf{(e)} \cite{babacan2011variational}}\medskip
	\end{minipage}
	\hfill
	\begin{minipage}[b]{0.2\linewidth}
		\centering
		\centerline{\includegraphics[width=3cm]{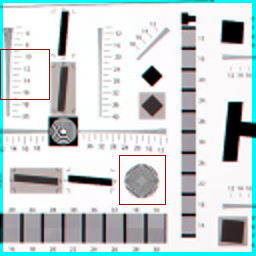}}
		\vspace{0.1cm}
		\includegraphics[width=1.1cm]{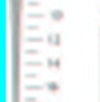}
		\includegraphics[width=1.1cm]{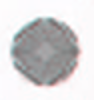}
		\centerline{\textbf{(f)} \cite{farsiu2004fast}}\medskip
	\end{minipage}
	\hfill
	\begin{minipage}[b]{0.2\linewidth}
		\centering
		\centerline{\includegraphics[width=3cm]{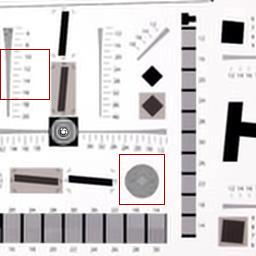}}
		\vspace{0.1cm}
		\includegraphics[width=1.1cm]{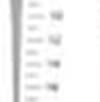}
		\includegraphics[width=1.1cm]{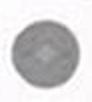}
		\centerline{\textbf{(g)} \cite{dong2011image}}\medskip
	\end{minipage}
	\hfill
	\begin{minipage}[b]{0.2\linewidth}
		\centering
		\centerline{\includegraphics[width=3cm]{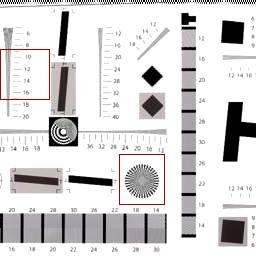}}
		\vspace{0.1cm}
		\includegraphics[width=1.1cm]{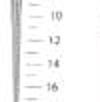}
		\includegraphics[width=1.1cm]{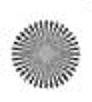}
		\centerline{\textbf{(h)} Ours}\medskip
	\end{minipage}
	\caption{Resolution chart image comparison ($\times 2$), with zoomed parts below reconstructed ones.}
	\label{fig:chart}
\end{figure*}

\begin{figure*}[h]
	\begin{minipage}[b]{0.2\linewidth}
		\centering
		\centerline{\includegraphics[width=3.2cm]{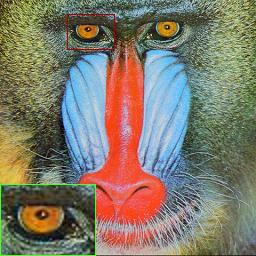}}
		\centerline{\textbf{(a)} Mandrill HR image}\medskip
	\end{minipage}
	\hfill
	\begin{minipage}[b]{0.2\linewidth}
		\centering
		\centerline{\includegraphics[width=3.2cm]{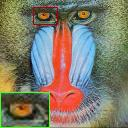}}
		\centerline{\textbf{(b)} LR reference image}\medskip
	\end{minipage}
	\hfill
	\begin{minipage}[b]{0.2\linewidth}
		\centering
		\centerline{\includegraphics[width=3.2cm]{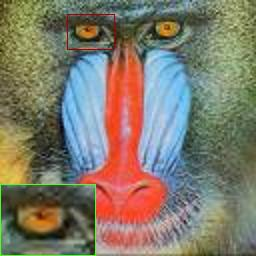}}
		\centerline{\textbf{(c)} Bicubic}\medskip
	\end{minipage}
	\hfill
	\begin{minipage}[b]{0.2\linewidth}
		\centering
		\centerline{\includegraphics[width=3.2cm]{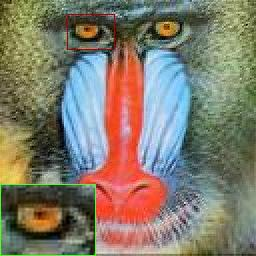}}
		\centerline{\textbf{(d)} \cite{zomet2001robust}}\medskip
	\end{minipage}
	\hfill
	\begin{minipage}[b]{0.2\linewidth}
		\centering
		\centerline{\includegraphics[width=3.2cm]{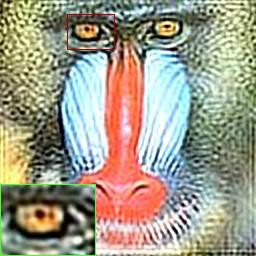}}
		\centerline{\textbf{(e)} \cite{babacan2011variational}}\medskip
	\end{minipage}
	\hfill
	\begin{minipage}[b]{0.2\linewidth}
		\centering
		\centerline{\includegraphics[width=3.2cm]{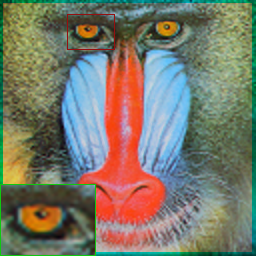}}
		\centerline{\textbf{(f)} \cite{farsiu2004fast}}\medskip
	\end{minipage}
	\hfill
	\begin{minipage}[b]{0.2\linewidth}
		\centering
		\centerline{\includegraphics[width=3.2cm]{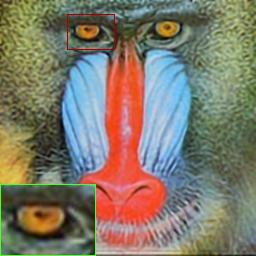}}
		\centerline{\textbf{(g)} \cite{dong2011image}}\medskip
	\end{minipage}
	\hfill
	\begin{minipage}[b]{0.2\linewidth}
		\centering
		\centerline{\includegraphics[width=3.2cm]{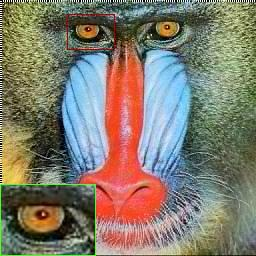}}
		\centerline{\textbf{(h)} Ours}\medskip
	\end{minipage}
	\caption{Mandrill image comparison ($\times 2$), with zoomed part in green.}
	\label{fig:mandrill}
\end{figure*}

\begin{figure*}[h]
	\centering
	\begin{minipage}[b]{0.2\linewidth}
		\centering
		\centerline{\includegraphics[width=3.2cm]{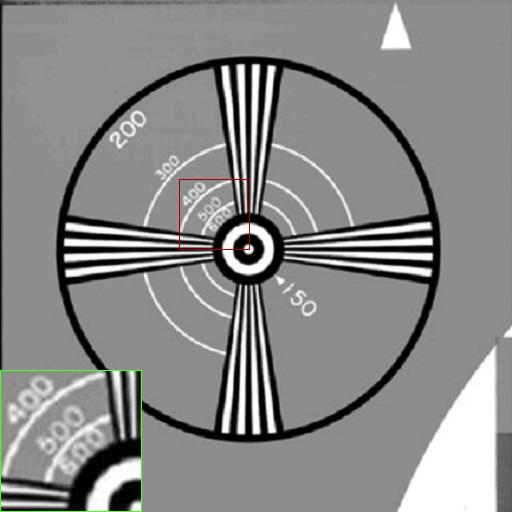}}
		\centerline{\textbf{(a)} Circles HR image}\medskip
	\end{minipage}
	\hfill
	\begin{minipage}[b]{0.2\linewidth}
		\centering
		\centerline{\includegraphics[width=3.2cm]{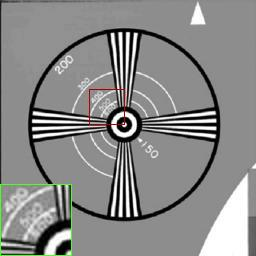}}
		\centerline{\textbf{(b)} LR reference image}\medskip
	\end{minipage}
	\hfill
	\begin{minipage}[b]{0.2\linewidth}
		\centering
		\centerline{\includegraphics[width=3.2cm]{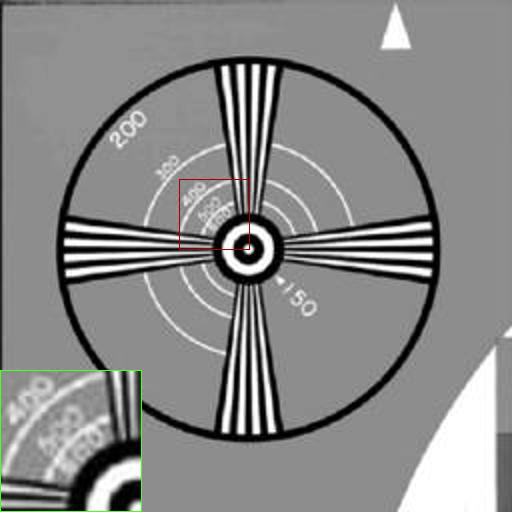}}
		\centerline{\textbf{(c)} Bicubic}\medskip
	\end{minipage}
	\hfill
	\begin{minipage}[b]{0.2\linewidth}
		\centering
		\centerline{\includegraphics[width=3.2cm]{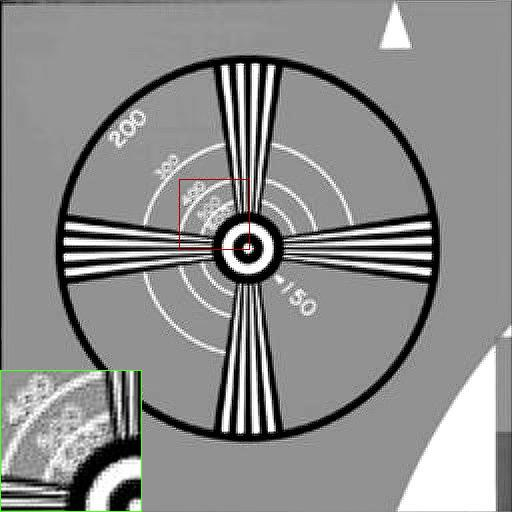}}
		\centerline{\textbf{(d)} \cite{zomet2001robust}}\medskip
	\end{minipage}
	\hfill
	\begin{minipage}[b]{0.2\linewidth}
		\centering
		\centerline{\includegraphics[width=3.2cm]{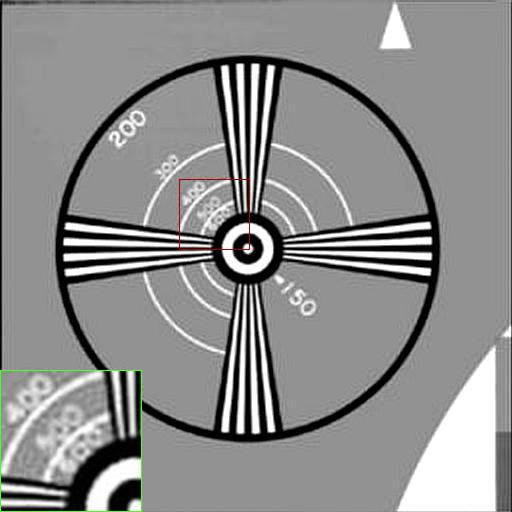}}
		\centerline{\textbf{(e)} \cite{babacan2011variational}}\medskip
	\end{minipage}
	\hfill
	\begin{minipage}[b]{0.2\linewidth}
		\centering
		\centerline{\includegraphics[width=3.2cm]{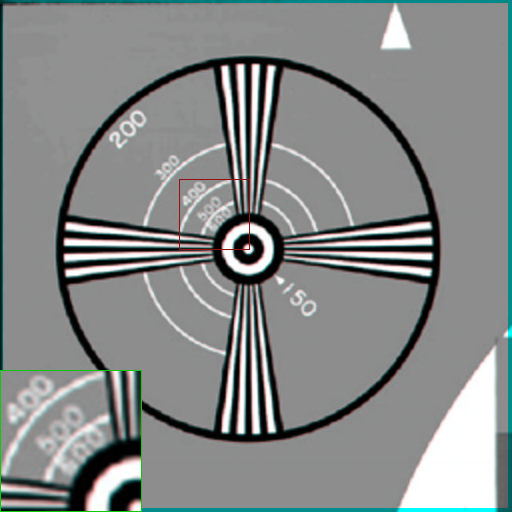}}
		\centerline{\textbf{(f)} \cite{farsiu2004fast}}\medskip
	\end{minipage}
	\hfill
	\begin{minipage}[b]{0.2\linewidth}
		\centering
		\centerline{\includegraphics[width=3.2cm]{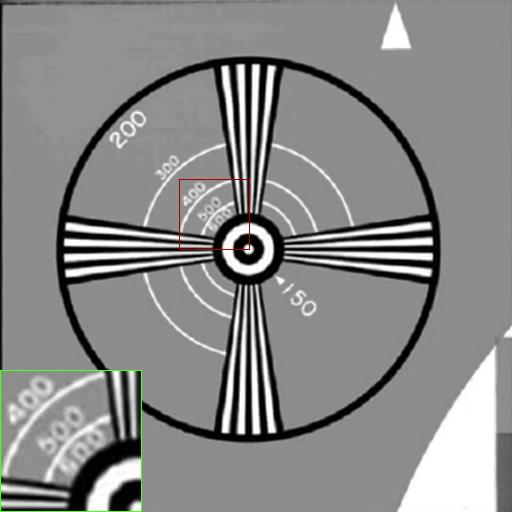}}
		\centerline{\textbf{(g)} \cite{dong2011image}}\medskip
	\end{minipage}
	\hfill
	\begin{minipage}[b]{0.2\linewidth}
		\centering
		\centerline{\includegraphics[width=3.2cm]{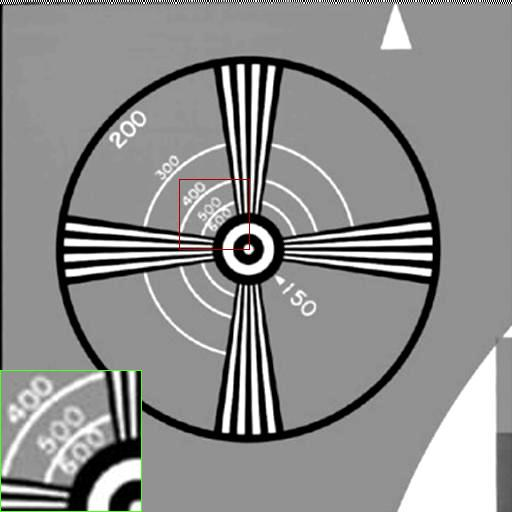}}
		\centerline{\textbf{(h)} Ours}\medskip
	\end{minipage}
	\caption{Circles (MDSP dataset \cite{farsiu2006multiframe}) image comparison ($\times 2$), with zoomed part in green.}
	\label{fig:circles}
\end{figure*}

\begin{figure*}[h]
	\centering
	\begin{minipage}[b]{0.2\linewidth}
		\centering
		\centerline{\includegraphics[width=3.2cm]{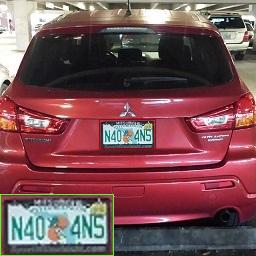}}
		\centerline{\textbf{(a)} Car HR image}\medskip
	\end{minipage}
	\hfill
	\begin{minipage}[b]{0.2\linewidth}
		\centering
		\centerline{\includegraphics[width=3.2cm]{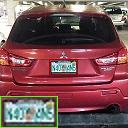}}
		\centerline{\textbf{(b)} LR reference image}\medskip
	\end{minipage}
	\hfill
	\begin{minipage}[b]{0.2\linewidth}
		\centering
		\centerline{\includegraphics[width=3.2cm]{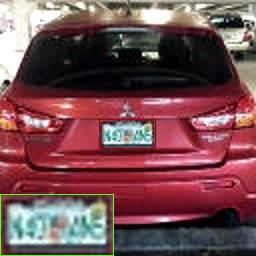}}
		\centerline{\textbf{(c)} Bicubic}\medskip
	\end{minipage}
	\hfill
	\begin{minipage}[b]{0.2\linewidth}
		\centering
		\centerline{\includegraphics[width=3.2cm]{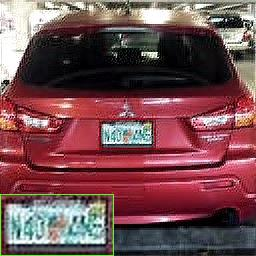}}
		\centerline{\textbf{(d)} \cite{zomet2001robust}}\medskip
	\end{minipage}
	\hfill
	\begin{minipage}[b]{0.2\linewidth}
		\centering
		\centerline{\includegraphics[width=3.2cm]{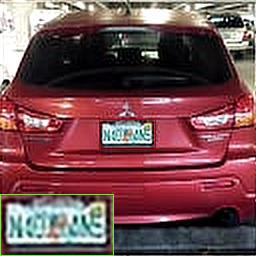}}
		\centerline{\textbf{(e)} \cite{babacan2011variational}}\medskip
	\end{minipage}
	\hfill
	\begin{minipage}[b]{0.2\linewidth}
		\centering
		\centerline{\includegraphics[width=3.2cm]{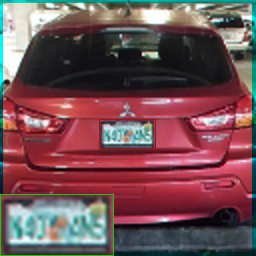}}
		\centerline{\textbf{(f)} \cite{farsiu2004fast}}\medskip
	\end{minipage}
	\hfill
	\begin{minipage}[b]{0.2\linewidth}
		\centering
		\centerline{\includegraphics[width=3.2cm]{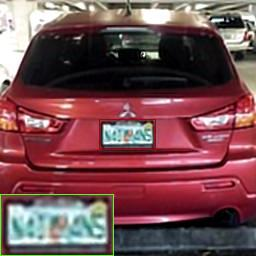}}
		\centerline{\textbf{(g)} \cite{dong2011image}}\medskip
	\end{minipage}
	\hfill
	\begin{minipage}[b]{0.2\linewidth}
		\centering
		\centerline{\includegraphics[width=3.2cm]{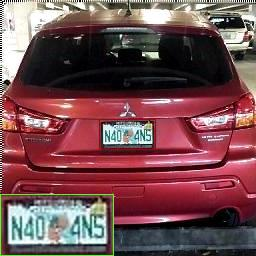}}
		\centerline{\textbf{(h)} Ours}\medskip
	\end{minipage}
	\caption{Car image comparison ($\times 2$), with zoomed part in green. Image is captured by a cell-phone camera.}
	\label{fig:car}
\end{figure*}

Figs. \ref{fig:lena}, \ref{fig:tag}, and \ref{fig:chart} to \ref{fig:car} show results obtained with our method, the state-of-the-art, and conventional ones. As can be seen from these figures, in zoomed areas particularly, the proposed method generates sharper edges with less artifacts compared to other methods. 

While bicubic interpolation ((c) parts in all related figures) tend to introduce blur to the images, Robust SR technique leaves jaggy artifacts on the edges which are easily seen in (d) parts of Figs. \ref{fig:lena}, \ref{fig:tag}, and \ref{fig:chart} of Lena, Car tag, and Resolution chart images. Babacan's method alleviates the jaggy artifacts in most cases, yet the final results remain overly smoothed; whereas, Farsiu's method does not reconstrut details that can be observed especially in part (f) of Figs. \ref{fig:chart} and \ref{fig:mandrill} of Resolution chart and Mandrill images. Dong's method, even though better at removing artifacts and achieving natural looking results compared to the other methods, also is prone to leave blurry images as can be recognized without difficulty in (g) parts of Resolution chart, Mandrill and Car images in Figs. \ref{fig:chart}, \ref{fig:mandrill}, and \ref{fig:car}. In addition, compared methods fail to recover fine details which is mostly recognized in car tags and numbers in Figs. \ref{fig:tag}, \ref{fig:chart}, \ref{fig:circles}, and \ref{fig:car}. 

On the other hand, the proposed method, seen in (h) parts of Figs. \ref{fig:lena}, \ref{fig:tag}, and \ref{fig:chart} to \ref{fig:car}, is able to recover sharp edges without visual artifacts, or blurring the images which leads to generating the closest results to the ground truth. Particularly, Resolution chart, Mandrill and Car images in Figs. \ref{fig:chart}, \ref{fig:mandrill}, and \ref{fig:car} demonstrate the high quality achieved with the proposed method. Recovered texture details with our method can be observed in all test cases upon a closer look, specifically in the feather texture of Lena's hat in Fig. \ref{fig:lena} and hair texture in cheeks of Mandrill image in Fig. \ref{fig:mandrill}. Overall, the proposed method removes artifacts and blur while preserving sharp edges without sacrificing a natural look. 

\subsection{Quantitative Comparison} 
To further investigate the effectiveness of our method, we also conduct a comparison based on objective measurements PSNR, RMSE and SSIM (\cite{wang2004image}), summarized in Table \ref{compPSNR}. Comparisons are based on the illuminance channel of images reconstructed with all methods. The best of all cases are in bold text. While Robust SR performs the worst based on most measurements for all images except Circles in Fig. \ref{fig:circles}, Dong's method has much better results compared to the other methods since in order to alleviate the correspondence ambiguity, their method uses different patches in a single image to learn various sets of bases. However, as can be seen also from Table \ref{compPSNR}, in most cases, quantitative comparisons confirm visual ones which shows that our method outperforms the state-of-the-art. Even though for Circles image in Fig. \ref{fig:circles}, the quantitative measurements are better for Dong's method, with a closer look in the green bordered square, it can be seen that details are recovered better in our method, e.g. "600" circle.

\begin{center}
	\begin{table*}[t]\setlength{\tabcolsep}{1.2pt}
		\centering
		\caption{Comparison of proposed method with other methods in PSNR, RMSE, SSIM.}
		\begin{tabular}{ r|rrr|rrr|rrr|rrr|rrr|rrr }
			\hline
			\multirow{2}{*}{Image} & \multicolumn{3}{c}{Bicubic} & \multicolumn{3}{c}{\cite{zomet2001robust}} & \multicolumn{3}{c}{\cite{babacan2011variational}} & \multicolumn{3}{c}{\cite{farsiu2004fast}} & \multicolumn{3}{c}{\cite{dong2011image}} & \multicolumn{3}{c}{Proposed} \\
			\cline{2-19}
			& PSNR & RMSE & SSIM & PSNR & RMSE & SSIM & PSNR & RMSE & SSIM & PSNR & RMSE & SSIM & PSNR & RMSE & SSIM & PSNR & RMSE & SSIM \\
			\cline{1-19}
			Lena & 26.57 & 11.97 & 0.81 & 22.69 & 18.70 & 0.68 & 22.51 & 19.1 & 0.77 & 26.82 & 11.62 & 0.83 & 28.37 & 9.73 & 0.88 & \textbf{32.65} & \textbf{5.94} & \textbf{0.96} \\
			Car tag & 26.86 & 11.58 & 0.87 & 18.57 & 30.06 & 0.54 & 22.92 & 18.21 & 0.78 & 23.83 & 17.38 & 0.84 & 29.86 & 8.19 & 0.93 & \textbf{33.15} & \textbf{5.61} & \textbf{0.97}\\
			Chart & 24.28 & 15.57 & 0.84 & 19.80 & 26.10 & 0.72 & 22.00 & 20.26 & 0.79 & 23.00 & 18.05 & 0.85 & 26.07 & 12.68 & 0.89 & \textbf{29.85} & \textbf{8.20} & \textbf{0.96} \\
			Mandrill & 22.13 & 19.94 & 0.61 & 20.76 & 23.37 & 0.50 & 15.15 & 44.59 & 0.42 & 22.98 & 18.09 & 0.63 & 22.25 & 19.69 & 0.63 & \textbf{24.09} & \textbf{15.93} & \textbf{0.90} \\
			Circles & 14.15 & 50.02 & 0.66 & 22.55 & 19.02 & 0.96 & 25.23 & 13.97 & 0.97 & 27.42 & 10.85 & 0.98 & \textbf{37.33} & \textbf{3.47} & \textbf{0.99} & 33.57 & 5.34 & \textbf{0.99}\\
			Car & 24.01 & 16.07 & 0.80 & 21.43 & 21.64 & 0.71 & 21.47 & 21.54 & 0.74 & 23.23 & 17.59 & 0.82 & 25.02 & 14.30 & 0.86 & \textbf{31.14} & \textbf{7.07} & \textbf{0.97} \\
			\bottomrule
		\end{tabular} \label{compPSNR}
	\end{table*}
\end{center}

\subsection{Computational Efficiency} The computational complexity of the proposed method depends on matrix multiplications ($\mathcal{O}(n^3)$) along with the TSVD method ($\mathcal{O}(mk^2)$ where $k$ is the approximation rank). Since all blocks have the same size and use the same shift information, matrix inversions are handled only once, and the proposed super resolution method is applied to all blocks in parallel. Our method can also be applied as a sparse method in order to reduce time complexity, considering the fact that coefficient matrices are either bidiagonal or at most triangular matrices. 

Time complexity of the proposed method and state-of-the-art is compared in Table \ref{compTime}, where average time taken for different size LR images is shown in seconds. Block size of all compared cases are set to $32 \times 32$.  As can be seen from the table, the proposed method outperforms regularization based method (\cite{babacan2011variational}) (where outliers of \cite{babacan2011variational} are removed for a fair comparison) and learning based method ({\cite{dong2011image}}) especially when the image sizes are relatively large. Since \cite{yang2010image} learns a compact representation for image patch pairs in LR and HR images to capture the cooccurrence prior, their method has a lower computational time complexity for smaller size images; however, as the size increases, the proposed method outperforms \cite{yang2010image} as well. 

\begin{figure}[h]
	\begin{minipage}[b]{0.5\linewidth}
		\centering
		\centerline{\includegraphics[width=8cm]{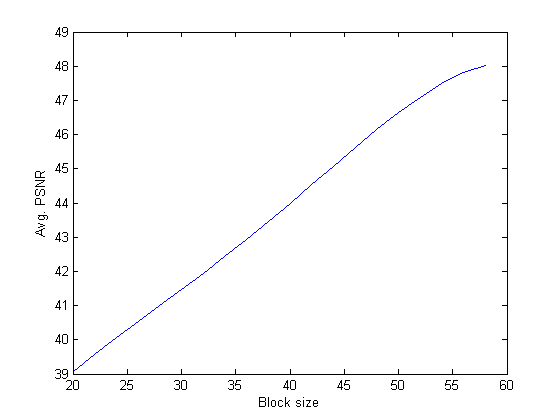}}
		\centerline{\textbf{(a)}}\medskip
	\end{minipage}
	\hfill
	\begin{minipage}[b]{0.5\linewidth}
		\centering
		\centerline{\includegraphics[width=8cm]{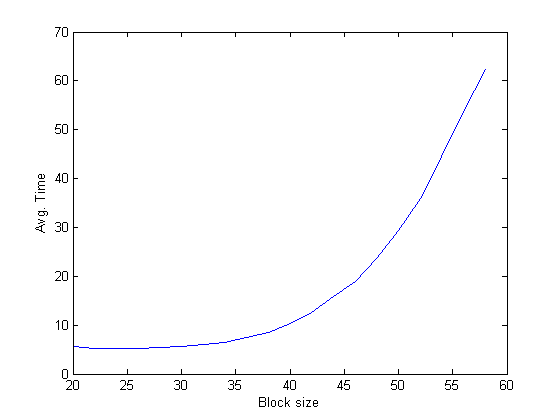}}
		\centerline{\textbf{(b)}}\medskip
	\end{minipage}
	\caption{\textbf{(a)} Average PSNR based on block size \textbf{(b)} Average runtime (\textit{s}) based on block size}\medskip \label{fig:psnrblock}
\end{figure}

A comparison of block sizes with time in seconds and PSNR for the proposed method is shown in Fig. \ref{fig:psnrblock}. The results are calculated for 100 different images for 100 random shift amounts, and the average time and PSNR are shown in the graphs (after removing the outliers). As one can see from the graph, as block sizes increase, PSNR improves; however, time complexity increases at the same time. Therefore, the block sizes can be decided based on the application depending on the importance of time or accuracy. Although the graph demonstrates the results for square sized blocks, the block sizes are decided based on the image sizes, which can as well be rectangular.

\begin{center}
	\begin{table*}[t]\setlength{\tabcolsep}{2pt}
		\centering
		\caption{Comparison of proposed method with other methods in time (s).}
		\begin{tabular}{r|rrrrrr}
			\toprule
			LR Image size & Bicubic & \cite{zomet2001robust} & \cite{babacan2011variational} & \cite{farsiu2004fast} & \cite{dong2011image} & Proposed \\
			\midrule
			32 $\times$ 32 & 0.3 & 9.7 & 19 & 1.2 & 12 & 7\\
			64 $\times$ 64& 0.3 & 9.6 & 36 & 1.2 & 33 & 12\\
			128 $\times$ 128& 0.3 & 9.8 & 217 & 1.2 & 140 & 35 \\
			256 $\times$ 256& 0.3 & 9.8 & 976 & 1.2 & 440 & 122 \\
			\bottomrule
		\end{tabular} \label{compTime}
	\end{table*} 
\end{center}

\section{Summary and Conclusions} \label{summary}
As a final remark, a direct wavelet-based super resolution technique is proposed in this paper by first deriving exact in-band relationships between two subpixel shifted images, then utilizing these relationships in a linear system form to reconstruct high frequency information of a low resolution reference image. Our results outperform the conventional as well as advanced recently published methods. We attribute this to the accuracy, well-posedness and the linearity of the equations derived in Section \ref{shifts} and the inherent local nature of wavelets, making them very effective in signal localization. In summary, we present herein a method for super-resolution by effectively estimating the high frequency information in the Haar domain, which in a sense is a hybrid approach between single image and multi-image methods, taking advantage of the best of both worlds.  


{
\bibliographystyle{plain}
\bibliography{foroosh,corresreg,IEEEabrv,refs}
}

\end{document}